\documentclass[11pt]{article}
\usepackage{times}
\usepackage{url}
\usepackage{latexsym}
\usepackage{graphicx}
\usepackage[T1]{fontenc}
\usepackage[utf8]{inputenc}
\usepackage{tabularx}
\usepackage[margin=3.4cm]{geometry}
\usepackage{titlesec}
\usepackage{sectsty}
\usepackage{footnote}
\usepackage{colortbl}
\usepackage{verbatim}

\usepackage{subfig}

\usepackage{multirow}
\usepackage{ulem}
\usepackage{lipsum}
\usepackage{amsmath}
\usepackage{hyperref}
\usepackage{float}
\usepackage{rotating}

\usepackage{authblk}

\usepackage{abstract}

\title{A Quantitative and Qualitative Analysis of \\ Suicide Ideation Detection using Deep Learning}

\author{Siqu Long}
\author{Rina Cabral}
\author{Josiah Poon}
\author{Soyeon Caren Han\thanks{Corresponding Author}}

\affil{University of Sydney, Sydney, NSW, Australia}
\date{}

\begin{document}
\maketitle
\begin{abstract}%125-150 words!!!!
\noindent\textit{For preventing youth suicide, social media platforms have received much attention from researchers. A few researches apply machine learning, or deep learning-based text classification approaches to classify social media posts containing suicidality risk. This paper replicated competitive social media-based suicidality detection/prediction models. We evaluated the feasibility of detecting suicidal ideation using multiple datasets and different state-of-the-art deep learning models, RNN-, CNN-, and Attention-based models. Using two suicidality evaluation datasets, we evaluated 28 combinations of 7 input embeddings with 4 commonly used deep learning models and 5 pretrained language models in quantitative and qualitative ways. Our replication study confirms that deep learning works well for social media-based suicidality detection in general, but it highly depends on the dataset's quality.}
\end{abstract}

\section{Introduction}
The World Health Organisation\footnote{http://apps.who.int/gho/data/node.sdg.3-4-viz-2?lang=en} reported that suicide is the second leading cause of death in the world for those aged 15 to 24 years and yet, despite 50 years of mental health research, the predictive ability for suicidality\footnote{https://www.health.gov.au/internet/publications/publishing.nsf/Content/mental-pubs-m-mhaust2-toc~mental-pubs-m-mhaust2-hig~mental-pubs-m-mhaust2-hig-sui} has not improved \cite{franklin2017risk}. Understanding how young individuals communicate and express their thoughts is the key to detecting such suicidality. 

Amongst adolescents and young adults, social media platforms are commonly used to express their thoughts and feelings. Luxton et al. (2012) \cite{luxton2012social} indicated that social media has a significant potential for use as a youth suicide prevention tool. Some researchers have examined social media platforms such as Twitter, Reddit, and Facebook, and have applied machine learning or deep learning-based text classification approaches to classify social media posts containing suicidality. Those attempts have been made to identify social media language patterns that express suicidality. O'Dea et al. (2015) \cite{o2015detecting} examined the level of suicidality for individual tweets by using term frequency and support vector machine (SVM). Burnap et al. (2015) \cite{burnap2015machine}  evaluated suicide-related tweet classification by comparing the results of around 300 features using different machine classifiers. Sawhney et al. (2018) \cite{sawhney2018exploring} and Tdesse et al. (2019) \cite{tadesse2019detection} employed RNN-based deep learning to conduct sentence classification for detecting suicidal ideation. Within this suicidality detection in the social media paradigm, most existing studies have three major limitations: The first limitation would be that the previous work only focused on a single data source, so it is not generalisable. This is due to privacy and anonymity concerns borne out of the social stigma associated with mental illness and suicide. Researchers in suicidality detection using social media reported that they collected and annotated the dataset themselves, and these datasets are rarely made available to the public. Secondly, there is no standard or perfect solution for selecting and preparing features that would represent how young individuals communicate and express suicidality in social media. Different psychologists have different ideas regarding suicidality resulting in subjective feature selection methods. Moreover, there are many suicidality prediction studies using different methods and combinations of natural language processing for feature engineering, but most studies only propose what works best for the dataset that they are using. There is also the problem of testing different combinations of these features as there are too many methods to try. Finally, very few attempts \cite{sawhney2018exploring} have been made to employ deep learning classifiers that identify text containing suicidality. Furthermore, most of these studies only include an overview of their framework on the published papers. They do not share the actual code or detailed evaluation setup (i.e. hyper-parameters), making it almost impossible to fully replicate their result or build up on top of their studies. This is why most studies are being repeated through using different feature selection methods and machine- or deep-learning algorithms.

Thus, our study aims to contribute to the social media-based suicidality detection/prediction literature by using multiple suicide-related social media datasets and different deep learning models, which are state-of-the-art in natural language processing and text classification. The main contributions of this research are summarised as follows:
\begin{enumerate}
\item To the best of our knowledge, no other studies have provided a comparative quantitative and qualitative analysis study to show an overview of how well their proposed social media-based suicidality prediction framework works on different datasets. We aim to contribute to the literature by using multiple datasets. 
\item The study will isolate the results gained from exclusively using textual features and compare the performance to other studies that integrates other attributes.
\item We encourage other researchers in building up on our study by not just providing an overview of our methodology in this paper but also by providing them with the code and detailed evaluation setup that was used to generate the evaluations presented in the paper.
\end{enumerate}

\section{Evaluation Dataset}
One of the foremost challenges in suicidality detection is the lack of publicly available datasets. This lacking is due to privacy and anonymity concerns borne out of the social stigma associated with mental illness and suicide. Researchers in suicidality detection using social media reported that they collected and annotated the dataset by themselves, and these datasets are rarely made available to the public. Motivated by the need to replicate a dataset, we reviewed and selected several literatures to identify their data collection and annotation processes. They tend not to manually develop a word list to represent suicidal language. Starting from Jashinsky et al. (2014) \cite{jashinsky2014tracking}, several researchers \cite{o2015detecting,burnap2015machine} extracted or adopted a lexicon of terms, words and phrases, by collecting anonymised data from known suicide web forums, blogs and micro-blogs.

\subsection{Dataset1: Suicide-related Tweets Dataset}
The first dataset is a replication of the dataset built through searching suicide-related keywords and phrases which are then annotated by the experts or clinicians \cite{o2015detecting,burnap2015machine}. We collected 1.6 million tweets monitored by Twitter API\cite{go2009twitter} and eliminated noise or unnecessary information such as usernames (i.e. @alecmgo).  With this collected Twitter dataset, a tweet would be included if it matched any of the following terms used by Jashinsky et al. (2014) \cite{jashinsky2014tracking} and O'Dea et al. (2015) \cite{o2015detecting}:  
"suicidal; suicide; kill myself; my suicide note; my suicide letter; end my life; never wake up; can't go on; not worth living; ready to jump; sleep forever; want to die; be dead; better off without me; better off dead; suicide plan; suicide pact; tired of living; don't want to be here; die alone; go to sleep forever". 

After the data collection and preprocessing, a dataset consisting of 660 tweets was manually annotated. Human annotation was used to determine the level of concern within the suicide-related tweets. O'Dea et al. (2015)\cite{o2015detecting}, with their five researchers (three mental health researchers and two computer scientists), proposed three suicidality levels with detailed definitions and specific instructions as follows:
1) \textbf{Strongly Concerning (SC)}: a convincing display of serious suicidal ideation; the author conveys a serious and personal desire to complete suicide, e.g., "I want to die" or "I want to kill myself" in contrast to "I might just kill myself" or "when you call me that name, it makes me want to kill myself"; suicide risk is not conditional on some event occurring, unless that event is a clear risk factor for suicide, e.g., bullying, substance use; the risk of suicide appears imminent, e.g., "I am going to kill myself" versus "If this happens, I will kill myself"; a suicide plan and/or previous attempts are disclosed; little evidence to suggest that the tweet is flippant, e.g., tweets with "lol" or other forms of downplaying are not necessarily flippant and may still be included in this category;
2) \textbf{Possibly Concerning (PC)}: the default category for all tweets; a tweet will only be removed from this category \textit{if and only if} it is classified as `strongly concerning' or `safe to ignore';
3) \textbf{Safe to Ignore (SI)}: no reasonable evidence to suggest that the risk of suicide is present.

In order to replicate this data annotation study, we recruited three human annotators (1 psychologist and 2 computer scientists) and asked them to classify these tweets  based on the above criteria. A satisfactory agreement between the annotators can be inferred from Table 1(a). In conclusion, 103 out of 660 tweets (15.61\%) were classified into "strongly concerning".

\begin{figure}
\centering
  \includegraphics[width=0.75\linewidth]{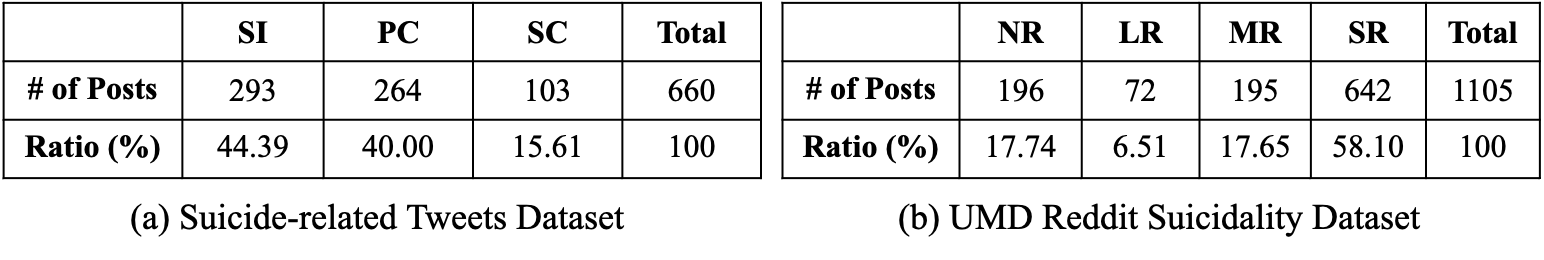}
  \caption{Data distribution of the two Suicidality Dataset: a)Suicide-related Tweets (SI: Safe to Ignore, PC: Possibly Concerning, SC: Strongly Concerning), b) UMD (NR: No Risk, LR: Low Risk, MR: Moderate Risk, SR: Severe Risk)}
  \label{fig:boat1}
\end{figure}

\begin{figure}[!ht]
\centering
 \subfloat[Suicide-related Tweets Dataset\label{fig:Twitter}]{%
   \includegraphics[width=0.35\textwidth]{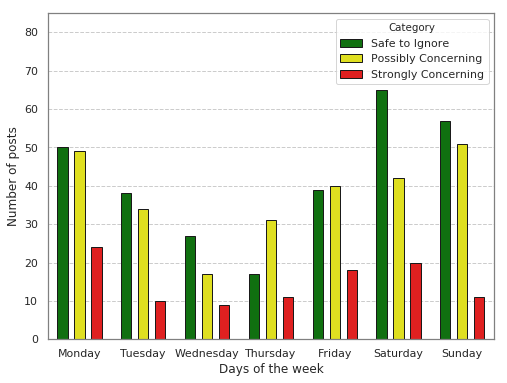}
 }
  \hspace{2.5em}
 \subfloat[UMD Reddit Suicidablity Dataset\label{fig:UMD}]{%
   \includegraphics[width=0.35\textwidth]{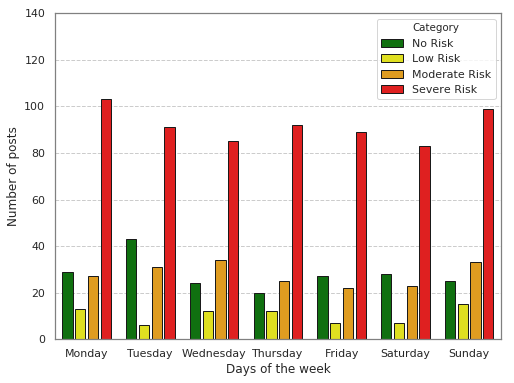}
 }
 \caption{Number of Posts in Each Day of the Week}
 \label{fig:dayofweek}
\end{figure}

\subsection{Dataset2: UMD Reddit Suicidality Dataset}
The second dataset used in this paper is the Reddit Suicidality Dataset collected by Shing et al. \cite{shing2018expert} of the University of Maryland for their study on suicide risk assessment via online postings. Instead of searching suicide-related posts based on a bag of predefined words or phrases, they assessed posts from users who were active on the Reddit suicide web forum, called "SuicideWatch". This dataset features a user-level classification where, instead of classifying each individual post, a user was classified based on all the posts they made on the forum. Since the goal of this study is to assess individual posts, user-level classes were carried over to each of their posts.  For instance, if the user was classified as a "severe risk" user, all posts published by the user would be annotated as "severe risk" level. The users and posts in the forum were assessed by experts and crowd annotators as either "no risk", "low risk", "moderate risk" and "severe risk". A total of 1,105 posts made by 496 users were gathered from the forum. There were two types of information included in the dataset: user and post details. User information contained an anonymised user ID and the annotated risk level for the user, while post information included post ID, user ID, timestamp, subreddit, post title and post body.  We received the data directly from the University of Maryland \cite{zirikly2019clpsych}. Table 2(b) shows the distribution of posts among the four suicidality levels. Since the dataset was collected from the suicide forum, it was not surprising that more than 50\% of posts were classified into the level "severe risk".

\subsection{Data Analysis}
This section presents the interesting points we found from the two datasets. We analysed the day of the week when each Twitter and Reddit post was published for both datasets. Figure \ref{fig:dayofweek} shows the temporal distribution of the posts used in the evaluation. Figure~\ref{fig:Twitter} shows~\textit{Dataset1: Suicide-related Tweets Dataset}, and Figure~\ref{fig:UMD} presents~\textit{Dataset2: UMD Reddit Suicidality Dataset}. The day distribution in both datasets shows that \textbf{"severe risk"} and \textbf{"strongly concerning"} users posted more on Mondays compared to the rest of the week. On the other hand, there was nothing special on Monday for the class \textbf{"safe to ignore"} in the Twitter dataset and the class \textbf{"no risk"} in the UMD.

\section{Experiment Setup}
\subsection{Baselines}
To validate the feasibility of deep learning models to detect suicidality in social media, baseline experiments were conducted. We reviewed literatures which focused on classifying text relating to suicide in social media and selected two papers which were published within last 5 years and has received more than 50 citations\footnote{Checked from https://scholar.google.com in March, 2022}.
\textbf{1) Baseline 1:} First, O'Dea et al. (2015) \cite{o2015detecting} applied SVM-based text classification for suicidality detection and used  TF-IDF score as a feature. This is the first paper that established the feasibility of detecting the level of concern for individual Twitter posts. \textbf{2) Baseline 2:} Burnap et al. (2015)  \cite{burnap2015machine} proposed a framework with three types of features derived from the Twitter posts: 1) features representing lexical characteristics of the sentences, 2) features representing sentiment, affective, and emotional features and level of terms used within the text, 3) features representing idiosyncratic language expression, and applied rotation forest. Those baselines are replicated based on their descriptions written in the literature.

\subsection{Feature Embeddings}
In Natural Language Processing with deep neural networks, the choice of an embedding approach influences the learning success of a network. In this evaluation, we concentrated on how to provide word-, character-, and document-level information to the classifiers as these types of information are most relevant and most frequently used for training deep learning-based text classification models \cite{horsmann2017lstms,sawhney2018exploring}. We also organised different network setups which have previously been used in different deep learning based suicidality detection in social media. 

%\paragraph{\textbf{Word}} 
\textbf{1)Word} We used context-independent word embeddings, GloVe \cite{pennington2014glove}, a global log-bilinear regression model for unsupervised learning of word representations that outperforms and is more efficient than Word2Vec \cite{mikolov2013distributed}. We used publicly available embeddings which were pre-trained on 2 billion English tweets (dimension size: 100). 

\textbf{2)Character}	We trained suicidality detection models with the character embedding model, Char2Vec \cite{cao2016joint}, which is an embedding associated with each character and is pre-trained on tweets and IMDB movie reviews (dimension size: 100). 

\textbf{3)Document}	We applied Doc2Vec \cite{le2014distributed}, a distributed representation of sentences and documents. The goal of Doc2Vec is to create a numeric representation of a document, regardless of its length. Doc2Vec uses the Word2Vec but added document-unique vector, paragraphID. This is because the documents do not come in logical structures such as words. We used publicly available pre-trained embedding on English Reddit posts and tweets (dimension size: 125).  

\textbf{4)Word+Char 5)Char+Doc 6)Word+Doc} These setups made use of a combination of 2 of the previous embeddings: word and character embeddings, character and document embeddings, word and document embeddings. For each word, these combination of embeddings were fed into different deep learning models for suicidality detection. 

\textbf{7)Word+Char+Doc} Lastly, we used all three embeddings for each word to train several deep learning-based suicidality detection models.

\subsection{Deep Learning Models}
Most text mining tasks have benefited greatly from the resurgence of deep neural networks. This section gives a brief introduction of the deep learning models we used for our experiments. We used Vanilla CNN, LSTM, Bi-LSTM, and Transformer.

\textbf{1)CNN} First, we used Vanilla Convolutional Neural Network (CNN)~\cite{cnn}. 

In this  setup, the vanilla CNN consists of three convolution layers, global max pooling, a dropout probability of 0.5 and a dense layer of 3 or 4 units depending on the number of classes of the dataset being evaluated. The convolution layers have 256, 128, and 64 hidden units respectively. Each layer has 3 as window size and uses ReLu for activation.

\textbf{2)LSTM} Recurrent Neural Network (RNN) has a fundamental issue of vanishing gradients so LSTM~\cite{hochreiter1997long} was proposed. LSTMs are well-suited to classify, process and predict time series and capture long-term dependencies in sentences along with a relative insensitivity to gap length. 

We applied Vanilla LSTM consisting of three LSTM layers with 128, 64 and 64 hidden units respectively followed by a dropout of 0.5 and 3 or 4 dense units depending on the number of classes of the dataset being evaluated.

\textbf{3)Bi-LSTM}	Bi-LSTM (Bi-directional LSTM)~\cite{bilstm} feeds the input in two ways, one from beginning to end and another from end to beginning. We applied three bi-directional LSTM layers of 64, 32 and 32 hidden units, a dropout of 0.5 and 3 or 4 dense units depending on the number of classes of the dataset being evaluated.

\textbf{4)Transformer}	Unlike RNN-based models, the transformer~\cite{vaswani2017attention} is an attention-only deep neural network that learns global-level information based on all the inputs and identifies the most important information. As the suicidality detection task did not require any sequential decoding process for each post, we adopted only the encoder mechanism of the transformer in order to predict the suicidality level of individual posts. Recently, transformer-based models achieve state-of-the-art results on multiple natural language processing tasks, especially text classification, so they motivated this study to explore it's performance for suicidality identification. In this setup, we applied a transformer with stack number L = 2, self-attention head number A = 5, feed-forward/filter size F = 400.

\subsection{Pretrained Language Model Architectures}
We also applied the transformer-based pretrained models, which are recently very successful in different NLP tasks. We applied \textbf{1)BERT~\cite{bert}, 2)RoBERTa~\cite{roberta}, 3)AlBERT~\cite{albert}, 4)ELECTRA~\cite{electra}, 5)GPT2~\cite{gpt2}}. For all the transformer-based pretrained language models, we fine-tuned the models on our suicide detection datasets using 2 hidden layers, 4 attention heads, ReLu hidden activation, and a 0.05 hidden dropout probability for 3 epochs each. For the Reddit dataset, we used a max length of 256.

\begin{figure}[!ht]
\captionsetup[subfigure]{justification=centering}
 \subfloat[Suicide-related Tweets Dataset \label{fig:Twittereval}]{%
   \includegraphics[width=0.45\textwidth]{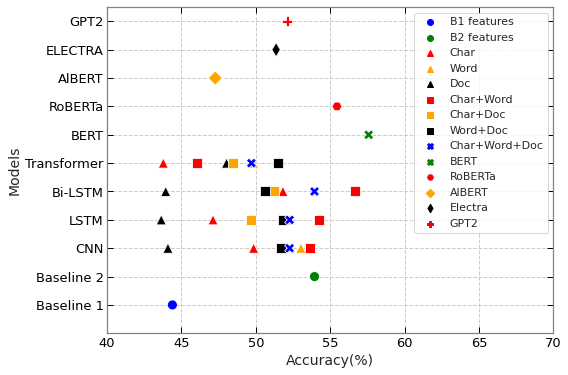}
 }
 \hfill
 \subfloat[UMD Reddit Suicidality Dataset
 \label{fig:UMDeval}]{%
   \includegraphics[width=0.45\textwidth]{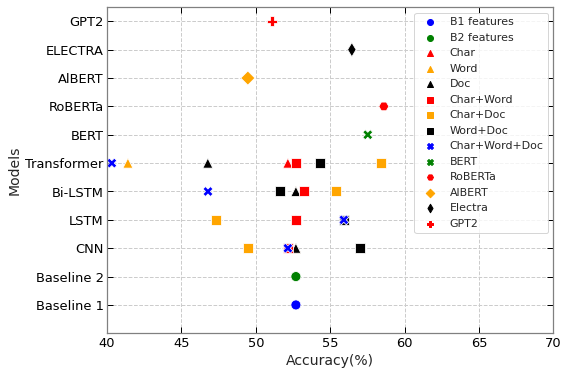}
 }
 \caption{Variance of Suicidality Detection in Social Media}
 \label{fig:eval}
\end{figure}

\section{Results}
In Figure \ref{fig:eval}, we show the suicidality classification accuracy of different embeddings and deep learning models for \textit{Dataset1: Suicide-related Tweets Dataset} and \textit{Dataset2: UMD Reddit Suicidality Dataset}\footnote{The detailed detection performance can be found in: \url{https://shorturl.at/inoFZ}}. The result represents different configurations of features and models for each dataset. First, we evaluated all permutations with 10-fold cross validation (10-fold CV) for the Dataset1 to be consistent with evaluations in other literatures using a similar type of dataset. As can be seen in the Figure \ref{fig:Twittereval} for Dataset1 that baseline 2 (with around 300 features rotation forest) achieves much higher performance than baseline 1 (with a TF-IDF feature and SVM), which is even competitive with other deep learning models. However, when having properly selected embedding types or being pretrained on large corpus, deep learning models can further lead to much better performance, such as Bi-LSTM with Char+Word embedding, RoBERTa or BERT (the best model). Besides, the three traditional deep learning models (RNN- or CNN-based models) illustrate similar taste of embedding types, e.g. doc embedding leads to the lowest while char+doc embedding generates the highest accuracy. For the second dataset \textit{Dataset2: UMD Reddit Suicidality Dataset}, we conducted 80-20 percentage split validation. This was the same training-testing split setup that was provided for the CLPsych 2019 Shared Task. In order to keep the original labels' distribution, they applied the proportional training/test split separately for each label. Figure \ref{fig:UMDeval} shows a very different pattern of results between the first and the second dataset. It is shown that baseline 1 becomes competitive and similar to baseline 2. The three deep learning models with varied embeddings result in different accuracy between the similar range of 45\%-60\%, most of which outperform the baseline models by a large margin. However, there is less observable pattern of embedding preference as in Dataset 1. The best performance was achieved by the pretrained RoBERTa model.

The different performance patterns can be explained by the characteristics of the two datasets. As mentioned in the section 3 'Evaluation Dataset', the first dataset \textit{Suicide-related Tweets Dataset} was gathered based on fixed suicide-related bag of keywords and phrases and were annotated based on the posts, while the posts (texts) in the second dataset \textit{UMD Reddit Suicidality Dataset} were annotated based on the overall risk of a user in the SuicideWatch forum. All posts published by 'medium risk' users were annotated as a 'medium risk' level, i.e. the post annotation process was not related to the actual text.

\begin{figure}[!ht]
\captionsetup[subfigure]{justification=centering}
\scriptsize
\begin{tabular}{p{7cm}  p{7cm}}
\textit{But there's allways people to help too i think i'd be dead now without kris or geraint or katie and
jess thanks you guys} & \textit{In the Munich Airport. Very classy first class lounge for Lufthansa (no, we're not flying FC). The
keg of Franziskaner seems to be dead} \\
\includegraphics[width=0.5\textwidth]{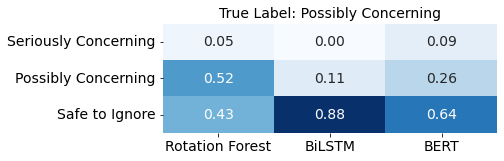} &
\includegraphics[width=0.45\textwidth]{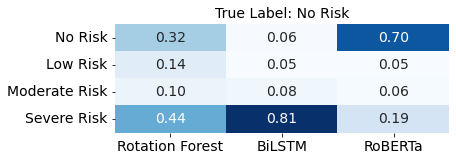} \\
\multicolumn{1}{c}{(a) Suicide-related Tweets Dataset}&\multicolumn{1}{c}{(b) UMD Reddit Suicidality Dataset}

\end{tabular}
 \caption{Case Study of Detailed Detection Probability Comparison on Suicide-related Tweets Dataset and UMD Reddit Suicidality Dataset}
 \label{fig:case_tweets}
\end{figure}

Figure~\ref{fig:case_tweets} illustrates the detailed suicide ideation detection result over the two randomly selected suicide posts (Dataset1 in Figure~\ref{fig:case_tweets}(a) and Dataset2 in Figure~\ref{fig:case_tweets}(b)), which compares the detailed prediction probability made by the following three model categories: baseline, traditional deep learning model and the pretrained language model. We use baseline 2 and Bi-LSTM as baseline model and traditional deep learning model representatives for both datasets considering their relatively stable and competitive performance while having BERT for Dataset1 and RoBERTa for Dataset2 as pretrained model representatives since they perform as the best models. Both Figure~\ref{fig:case_tweets}(a) and (b) display the text content of posts on the top while showing the probability heatmaps (extracted from the softmax layer output before classification) with the true labels in the bottom. It can be seen from Figure~\ref{fig:case_tweets}(a) that the post from Dataset1 that, when the data is annotated based on the post and easy to understand properly purely based on the textual words, Rotation Forest tends to classify correctly as `possibly concerning'. In comparison, as can be seen from Dataset2 in Figure~\ref{fig:case_tweets}(b), solely relying on the presented text failed the Rotation Forest and Bi-LSTM while RoBERTa can still correctly classify the post as `no risk' even when the suicide-oriented word such as `dead' appears in the text, possibly due to its pre-learned generic contextualized knowledge that can understand the `Franziskaner' is a brand of beer.

\section{Conclusion}
In this paper, we replicated several competitive social media-based suicidality detection models and evaluated the feasibility of detecting suicidal ideation using multiple datasets and different state-of-the-art deep learning models namely RNN-, CNN-, and Attention-based models. We tested 28 combinations of 7 types of input embeddings with 4 commonly used deep learning models and 5 pretrained language models using two suicidality evaluation datasets. Our replication study confirmed the result in general, and we additionally found that attention-based deep learning has the advantage of being the best performer with large datasets. However, we also found that deep learning models do not work well in the example of very small suicidality data but it would perform well if there is enough data available in the specific class. The paper is the first study that provided a comparative study to show an overview of how well other proposed social media-based suicidality prediction frameworks work on different datasets by replicating competitive existing studies and evaluating deep learning models with different setups.

\bibliography{template-references.bib}

\begin{thebibliography}{10}

\bibitem{franklin2017risk}
Franklin JC, Ribeiro JD, Fox KR, Bentley KH, Kleiman EM, Huang X, et~al.
\newblock Risk factors for suicidal thoughts and behaviors: a meta-analysis of
  50 years of research.
\newblock Psychological Bulletin. 2017;143(2):187.

\bibitem{luxton2012social}
Luxton DD, June JD, Fairall JM.
\newblock Social media and suicide: a public health perspective.
\newblock American journal of public health. 2012;102(S2):S195-200.

\bibitem{o2015detecting}
O'Dea B, Wan S, Batterham PJ, Calear AL, Paris C, Christensen H.
\newblock Detecting suicidality on Twitter.
\newblock Internet Interventions. 2015;2(2):183-8.

\bibitem{burnap2015machine}
Burnap P, Colombo W, Scourfield J.
\newblock Machine classification and analysis of suicide-related communication
  on twitter.
\newblock In: Proceedings of the 26th ACM conference on hypertext \& social
  media. ACM; 2015. p. 75-84.

\bibitem{sawhney2018exploring}
Sawhney R, Manchanda P, Mathur P, Shah R, Singh R.
\newblock Exploring and learning suicidal ideation connotations on social media
  with deep learning.
\newblock In: Proceedings of the 9th Workshop on Computational Approaches to
  Subjectivity, Sentiment and Social Media Analysis; 2018. p. 167-75.

\bibitem{tadesse2019detection}
Tadesse MM, Lin H, Xu B, Yang L.
\newblock Detection of suicide ideation in social media forums using deep
  learning.
\newblock Algorithms. 2019;13(1):7.

\bibitem{jashinsky2014tracking}
Jashinsky J, Burton SH, Hanson CL, West J, Giraud-Carrier C, Barnes MD, et~al.
\newblock Tracking suicide risk factors through Twitter in the US.
\newblock Crisis. 2014.

\bibitem{go2009twitter}
Go A, Bhayani R, Huang L.
\newblock Twitter sentiment classification using distant supervision.
\newblock CS224N Project Report, Stanford. 2009;1(12):2009.

\bibitem{shing2018expert}
Shing HC, Nair S, Zirikly A, Friedenberg M, Daum{\'e}~III H, Resnik P.
\newblock Expert, crowdsourced, and machine assessment of suicide risk via
  online postings.
\newblock In: Proceedings of the Fifth Workshop on Computational Linguistics
  and Clinical Psychology: From Keyboard to Clinic; 2018. p. 25-36.

\bibitem{zirikly2019clpsych}
Zirikly A, Resnik P, Uzuner O, Hollingshead K.
\newblock CLPsych 2019 shared task: Predicting the degree of suicide risk in
  Reddit posts.
\newblock In: Proceedings of the Sixth Workshop on Computational Linguistics
  and Clinical Psychology; 2019. p. 24-33.

\bibitem{horsmann2017lstms}
Horsmann T, Zesch T.
\newblock Do LSTMs really work so well for PoS tagging?--A replication study.
\newblock In: Proceedings of the 2017 Conference on Empirical Methods in
  Natural Language Processing; 2017. p. 727-36.

\bibitem{pennington2014glove}
Pennington J, Socher R, Manning C.
\newblock Glove: Global vectors for word representation.
\newblock In: Proceedings of the 2014 conference on empirical methods in
  natural language processing (EMNLP); 2014. p. 1532-43.

\bibitem{mikolov2013distributed}
Mikolov T, Sutskever I, Chen K, Corrado GS, Dean J.
\newblock Distributed representations of words and phrases and their
  compositionality.
\newblock In: Advances in neural information processing systems; 2013. p.
  3111-9.

\bibitem{cao2016joint}
Cao K, Rei M.
\newblock A Joint Model for Word Embedding and Word Morphology.
\newblock ACL 2016. 2016:18.

\bibitem{le2014distributed}
Le Q, Mikolov T.
\newblock Distributed representations of sentences and documents.
\newblock In: International conference on machine learning; 2014. p. 1188-96.

\bibitem{cnn}
Krizhevsky A, Sutskever I, Hinton GE.
\newblock ImageNet Classification with Deep Convolutional Neural Networks.
\newblock In: Pereira F, Burges CJC, Bottou L, Weinberger KQ, editors. Advances
  in Neural Information Processing Systems. vol.~25. Curran Associates, Inc.;
  2012. Available from:
  \url{https://proceedings.neurips.cc/paper/2012/file/c399862d3b9d6b76c8436e924a68c45b-Paper.pdf}.

\bibitem{hochreiter1997long}
Hochreiter S, Schmidhuber J.
\newblock Long short-term memory.
\newblock Neural computation. 1997;9(8):1735-80.

\bibitem{bilstm}
Schuster M, Paliwal KK.
\newblock Bidirectional recurrent neural networks.
\newblock IEEE transactions on Signal Processing. 1997;45(11):2673-81.

\bibitem{vaswani2017attention}
Vaswani A, Shazeer N, Parmar N, Uszkoreit J, Jones L, Gomez AN, et~al.
\newblock Attention is all you need.
\newblock Advances in neural information processing systems. 2017;30.

\bibitem{bert}
Kenton JDMWC, Toutanova LK.
\newblock BERT: Pre-training of Deep Bidirectional Transformers for Language
  Understanding.
\newblock In: Proceedings of NAACL-HLT; 2019. p. 4171-86.

\bibitem{roberta}
Liu Y, Ott M, Goyal N, Du J, Joshi M, Chen D, et~al.
\newblock RoBERTa: A Robustly Optimized BERT Pretraining Approach.
\newblock arXiv e-prints, pparXiv-1907. 2019.

\bibitem{albert}
Lan Z, Chen M, Goodman S, Gimpel K, Sharma P, Soricut R.
\newblock ALBERT: A Lite BERT for Self-supervised Learning of Language
  Representations.
\newblock Proceedings of NAACL-HLT. 2020.

\bibitem{electra}
Clark K, Luong MT, Le QV, Manning CD.
\newblock ELECTRA: Pre-training Text Encoders as Discriminators Rather Than
  Generators.
\newblock arXiv preprint arXiv:200310555. 2020.

\bibitem{gpt2}
Radford A, Wu J, Child R, Luan D, Amodei D, Sutskever I, et~al.
\newblock Language models are unsupervised multitask learners.
\newblock OpenAI blog. 2019;1(8):9.

\end{thebibliography}
\end{document}